\documentclass[11pt]{article}

\usepackage[final]{acl}

\usepackage{times}
\usepackage{latexsym}

\usepackage[T1]{fontenc}
\usepackage[utf8]{inputenc}

\usepackage{microtype}

\usepackage{inconsolata}

\usepackage{graphicx}

\usepackage{graphicx}
\usepackage{microtype}
\usepackage{amsmath}
\usepackage{amssymb}
\usepackage{dsfont}
\usepackage{booktabs}
\usepackage{algorithm}
\usepackage{algorithmic}
\usepackage[most]{tcolorbox}
\usepackage{enumitem}
\tcbuselibrary{skins}
\usepackage{listings}
\usepackage{xcolor}
\usepackage{subcaption}
\usepackage{multirow, multicol}
\usepackage{adjustbox}
\lstdefinestyle{promptstyle}{
  basicstyle=\ttfamily\small,
  columns=fullflexible,
  breaklines=true,
  frame=none
}
\usepackage{mdframed}
\usepackage{stfloats}
\definecolor{darkblue}{rgb}{0, 0, 0.5}
\hypersetup{colorlinks=true, citecolor=darkblue, linkcolor=darkblue, urlcolor=darkblue}

\newtcolorbox{notebox}[1][Note]{
  colback=blue!5!white,
  colframe=blue!75!black,
  fonttitle=\bfseries,
  title={#1},
  boxrule=0.5pt,
  arc=2pt,
}

\lstdefinestyle{pseudocode}{
  basicstyle=\ttfamily\footnotesize,   % smaller font
  keywordstyle=\bfseries,
  commentstyle=\color{gray}\itshape,
  showstringspaces=false,
  columns=fullflexible,                % tighter char spacing
  keepspaces=true,
  tabsize=2,                           % narrower indentation
  frame=single,
  numbers=left,
  numberstyle=\scriptsize\color{gray},
  numbersep=5pt,
  xleftmargin=1.2em,
  framexleftmargin=1.2em,
  breaklines=true,                     % wrap long lines
  breakatwhitespace=true,
  morekeywords={def,for,in,if,return,continue,nil},
  literate={<-}{{$\leftarrow$}}2 {||}{{$\Vert$}}1 {|_ord}{{$\cup_{\text{ord}}$}}3 {!=}{{$\neq$}}1,
}

\usepackage{tcolorbox}
\usepackage{listings}

\newtcolorbox{configbox}[1][]{
  colback=gray!10,
  colframe=black,
  fonttitle=\bfseries,
  title=#1,
  boxrule=1pt,
  arc=2pt,
  left=6pt,
  right=6pt,
  top=4pt,
  bottom=4pt
}

\usepackage{float}

\title{RAFT: A Stateful Retrieval-Augmented Framework for Troubleshooting Agents\thanks{Accepted to EMNLP 2026 (Industry Track)}}

\author{
  \textbf{Mingxuan Zhang}, \textbf{Xiaowen Wang}, \textbf{Anupma Sharan}, \textbf{Zhengyi Chen},
\\
  \textbf{Chenyu Diana Zhang}, \textbf{Shanshan Yang}, \textbf{Chittibabu Pacharu}
\\
\\
  Microsoft, Redmond, WA, USA
\\
  \small{\textbf{Correspondence:} \href{mailto:mingxzhang@microsoft.com}{mingxzhang@microsoft.com}}
}

\begin{document}
\maketitle
\begin{abstract}
Effective troubleshooting agents in enterprise customer support depend on retrieving actionable guidance from similar historical cases, yet existing retrieval-augmented generation (RAG) systems treat support cases as static documents and overlook their multi-stage, stateful nature. We introduce RAFT (Retrieval-Augmented Framework for Troubleshooting Agents), a stateful RAG framework that abstracts each closed historical case into a directed chain of timeline entries and retrieves at the entry level, surfacing cases whose intermediate states match the active case and returning the parent-case trajectory anchored at the matched state; an optional case-level graph links cases through a configurable similarity representation. We evaluate this retrieval layer directly, which, unlike evaluating a full agent system, requires no production deployment. Because public multi-stage troubleshooting data is extremely rare, we pair a synthetic benchmark built from Microsoft Learn Windows Server documentation with real Apache Jira issues carrying human-created duplicate labels. RAFT improves Case Hit over vanilla RAG and GraphRAG baselines at every stage of case progress, with statistically significant gains over the strongest baseline; the Jira results provide directional evidence that the advantage transfers to real case histories. We release our benchmark, implementation, and the Apache Jira evaluation set.\footnote{\url{https://github.com/microsoft/RAFT}}
\end{abstract}
 
\section{Introduction}
\label{sec:introduction}
 
Large language models (LLMs) now power intelligent agents across domains from software engineering \citep{jimenez2023swe} to customer support \citep{xu2024retrieval}. In enterprise customer support, an effective agent must reason over a private corpus of closed historical cases, whether it resolves incoming tickets or assists support engineers in doing so. Fine-tuning \citep{hu2022lora, ouyang2022training} can inject such domain knowledge, but it is computationally expensive, restricted to open-weight models, prone to catastrophic forgetting \citep{luo2025empirical}, and requires periodic retraining as new and more recent cases emerge. Retrieval-augmented generation (RAG)~\citep{lewis2020retrieval, gao2023retrieval, singh2025agentic} is a more practical alternative, grounding the LLM in a private knowledge base at inference time without altering its parameters. Yet as knowledge bases grow in scale and complexity, traditional RAG struggles: retrieved context is often extensive, poorly organized, and noisy, degrading both retrieval accuracy and the agent's ability to reason over it~\citep{han2024retrieval, edge2024local, xiang2025use, chen2024benchmarking}.
 
GraphRAG~\citep{edge2024local, zhang2025survey, zhuang2025linearrag, chen2025you, yang2026graph} responds by imposing explicit relational structure over the knowledge base. However, these general-purpose pipelines are not designed for troubleshooting histories, where investigations unfold across heterogeneous, noisy artifacts and closed cases vary in the actionable guidance they provide. Effective retrieval must identify relevant investigation states, preserve coherent case trajectories, distinguish useful evidence from non-actionable records, and protect sensitive information (Section~\ref{sec:problem_statement}). By contrast, entity-centric GraphRAG approaches build graphs of LLM-extracted entities and relations offline and use them to guide retrieval, without explicitly representing the progression of individual investigations. This adds graph-construction cost and couples retrieval to an entity-centric representation that can be harder to adapt as models and agent harnesses evolve~\citep{zhuang2025linearrag, xiang2025use, chen2025you}.

We propose RAFT (Retrieval-Augmented Framework for Troubleshooting Agents), a stateful RAG framework built around the multi-stage nature of troubleshooting. RAFT abstracts each closed historical case into a directed chain of timeline entries, each recording the technical state at one meaningful stage of the investigation. Embedding and retrieving at the entry level rather than the case level surfaces cases whose intermediate states most closely match the active case, giving the agent both tactical guidance for the current stage and strategic context on where comparable cases led. A complementary, optional case-level graph links cases through a configurable similarity representation, enabling expansion to relevant cases that do not match at the entry level.

To assess whether this representation supplies useful evidence throughout an investigation, we evaluate the retrieval layer independently of a complete troubleshooting agent. Realistic end-to-end evaluation often depends on access to operational environments and production workflows, making reproducible academic evaluation, cross-system comparison, and extension by others difficult. The retrieval layer, however, is independently testable across agent harnesses, underlying models, and workflows. We therefore measure whether, given the current information in an active case, RAFT surfaces similar historical cases that provide concrete evidence for diagnosis and resolution.

This evaluation requires multi-stage troubleshooting histories and labels identifying similar cases, but suitable public data is extremely rare (Section~\ref{sec:related_work}). We evaluate on two complementary datasets: a synthetic benchmark constructed from Microsoft Learn Windows Server troubleshooting documentation~\citep{MicrosoftDocs2024SupportArticles} for controlled demonstration and development, and real Apache Jira \citep{ApacheJira} issues with human-created duplicate labels to check that the advantage transfers to real data. RAFT improves Case Hit over vanilla RAG and GraphRAG baselines at every stage of case progress (Sections~\ref{sec:results} and~\ref{sec:jira}).

\paragraph{Contributions.}
\begin{itemize}[leftmargin=1.5em, itemsep=2pt, topsep=3pt, parsep=0pt, partopsep=0pt]
    \item A \textbf{stateful RAG architecture} that represents closed historical cases as directed chains of timeline entries, retrieves over their evolving intermediate troubleshooting states, and returns the parent-case trajectory anchored at the matched state.
    \item A \textbf{case-level graph} linking cases through configurable case attributes, enabling principled expansion beyond initial entry-level matches.
    \item A \textbf{public synthetic development benchmark} of 826 support cases from Microsoft Learn Windows Server documentation, with a reproducible protocol that probes retrieval at multiple stages of an active case.
\end{itemize}
 
\section{Related Work}
\label{sec:related_work}
 
For broad surveys of RAG, Agentic RAG, and GraphRAG, we refer readers to \citet{fan2024survey, zhang2025survey, singh2025agentic}. These pipelines target generic document corpora, and applying them to historical customer support cases remains underexplored, largely due to the lack of suitable public datasets. To our knowledge, no public customer support dataset~\citep{abdellatif2025helpdesktickets, qu2018analyzing, yang2018response} satisfies all of the following criteria: (1) rich, multi-turn interactions rather than single-turn question answering; (2) key entities (error codes, products, services) preserved without redaction; and (3) labels grouping similar cases together. Beyond the data gap, most work in this area originates in industry settings where evaluation relies on proprietary data and production metrics, severely limiting reproducibility. Existing academic work also focuses largely on question answering over static domain documents~\citep{su2025llm, patel2025graph, zhao2025agent} rather than leveraging historical case interactions for multi-stage troubleshooting. The closest work to ours, \citet{xu2024retrieval}, constructs a knowledge graph from historical cases for question answering and shows that graph structure can benefit support retrieval; however, their system is closed-source, their dataset private, and their evaluation confined to a production environment, making direct comparison infeasible. Evaluating the retrieval layer directly removes this dependence on production deployments and lets others reproduce and extend the comparison. We therefore pair a public synthetic benchmark for controlled demonstration and development with a transfer evaluation on Apache Jira~\citep{ApacheJira}, alongside an open implementation and a reproducible protocol.

\section{Problem Statement}
\label{sec:problem_statement}
 
In enterprise customer support, particularly for IT services, resolving a ticket is rarely a one-shot process. Resolution unfolds across multiple stages: an initial symptom report, hypothesis formation and iterative information gathering across logs, system outputs, and diagnostic tools, and finally identifying the root cause and issuing targeted remediation. Across this trajectory, surfacing similar previously resolved cases and learning how they were diagnosed and fixed is critical for an agent to resolve the active case effectively.
 
We therefore seek a retrieval system, built on a corpus of closed historical cases, that identifies similar cases given a query reflecting the current state of an active case and surfaces guidance on both the immediate next step and the broader resolution trajectory. Critically, retrieval must be \textbf{state-aware}: rather than treating the active case as a static lookup, the agent issues a sequence of updated queries as the active case evolves, and the system must return the most pertinent cases at each stage. Early on, a ticket describes only surface symptoms (an error code or a brief account of the behavior), and the agent should retrieve cases with similar initial symptoms to identify productive directions; as more context is gathered, updated queries let the system surface cases whose intermediate states match the current investigative state, giving finer-grained insight into the next steps.
 
We deliberately scope this retrieval problem to surfacing similar cases and the evidence needed to diagnose and act on them. Higher-order reasoning, over how cases relate or how root causes cluster, is valuable but is best performed by the agent online against the specific active case rather than frozen into the offline index. We therefore keep retrieval focused on this task and defer relational inference to the agent, a stance our metrics (Section~\ref{experiments}) reflect.
 
This problem exposes four fundamental limitations of existing RAG approaches:
 
\begin{enumerate}[leftmargin=1.5em, itemsep=3pt, topsep=3pt, parsep=0pt, partopsep=0pt]
    \item \textbf{Noisy, unstructured data.} Raw case data interleaves substantial non-technical content with the diagnostically relevant details, and the signal is scattered across many turns. Because existing RAG baselines operate directly on this raw data, the noise degrades their ability to identify genuinely similar cases in the first place.
    \item \textbf{Incoherent retrieved chunks.} Even when a similar case is retrieved, an agent cannot reason effectively from isolated chunks. To judge whether a historical case is truly similar and to learn from how it was resolved, the agent must see the case as a coherent whole, which requires reassembling the raw case data by mapping retrieved chunks back to their parent cases. Given the size of real cases and the practical output-token limits of RAG tools, this both wastes tokens and sharply limits the number of similar cases an agent can examine per retrieval. GraphRAG methods that return chunks together with their entity relationships partially mitigate this, but the construction process is unstable and offers no guarantee that the chunks needed to form a coherent picture are retained \citep{zhuang2025linearrag, han2025rag, zhou2025depth, xiang2025use}.
    \item \textbf{Non-actionable cases.} Closed tickets do not uniformly contain useful diagnostic or remediation evidence. Some end because the customer becomes unresponsive or the ticket is administratively closed, without documenting a meaningful investigation or outcome. A retrieval system must distinguish such records from cases that can inform the active investigation, rather than treating closure itself as evidence of usefulness.
    \item \textbf{Privacy constraints.} Enterprise customer data often contains personally identifiable information that should not be exposed directly to retrieval or to the agent, requiring an additional processing layer to abstract or redact such content before indexing.
\end{enumerate}
 
RAFT addresses these limitations through the two-level architecture described next.
 
\section{RAFT}
\label{sec:raft}
 
RAFT organizes closed historical cases at two levels: a per-case extracted resolution trajectory and a case-level graph $\mathcal{G}$ that links cases through a configurable similarity representation.
 
\begin{figure*}[t]
  \centering
  \includegraphics[width=\textwidth]{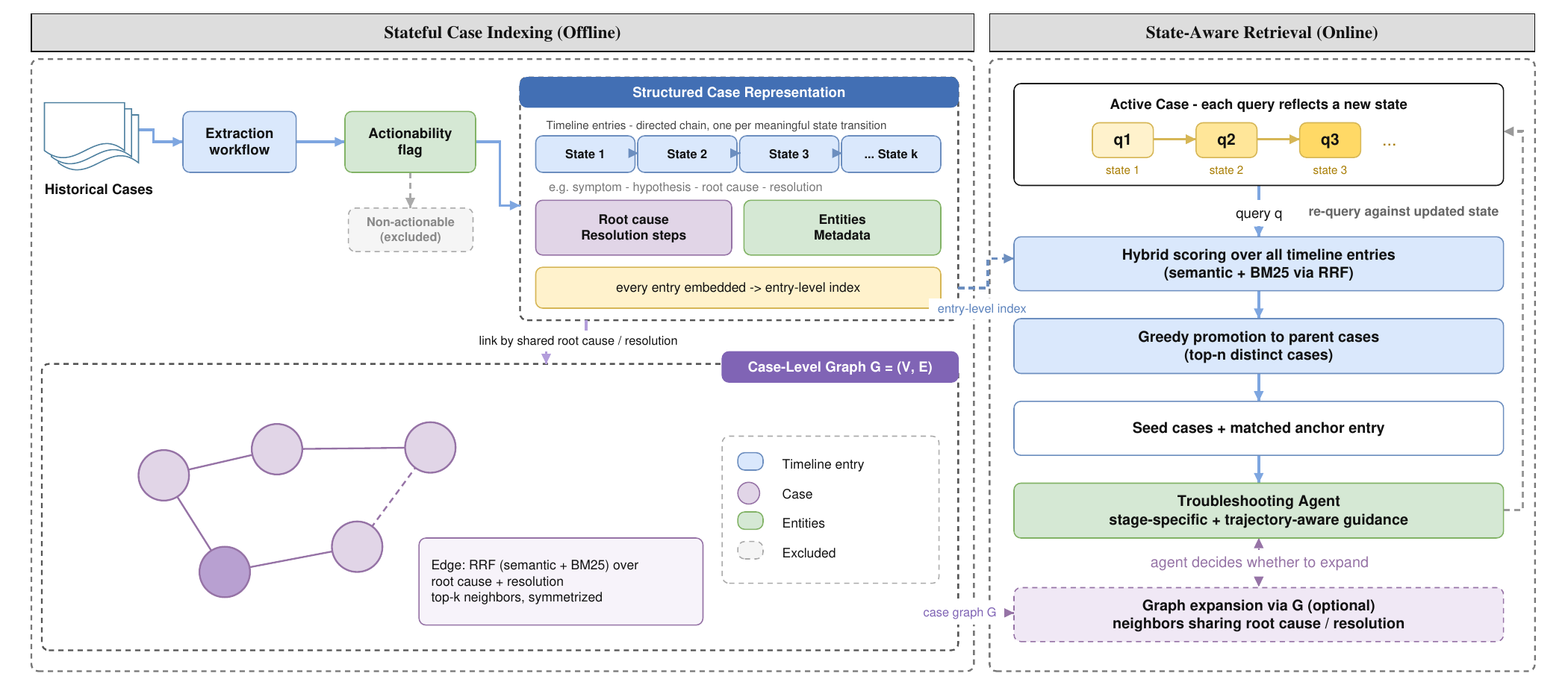}
  \caption{Overview of \textbf{RAFT}: workflow-based case extraction and entry-level retrieval with optional graph expansion. The actionability flag illustrates reviewer-based filtering; the feedback arrow denotes re-querying by the external troubleshooting agent.}
  \label{fig:raft-overview}
\end{figure*}
 
\subsection{Indexing}
 
Let $\mathcal{H} = \{h_i\}_{i=1}^{N}$ be a corpus of $N$ closed historical cases. Each raw case $h_i$ has a unique identifier $u_i$ (e.g., ticket number), metadata $m_i$ (e.g., category, created and closed times), and a time-ordered turn sequence $(x_1^{(i)}, \ldots, x_{T_i}^{(i)})$ of emails, notes, logs, and similar artifacts. We process each case independently to obtain a structured representation $\tilde{h}_i$:
 
\begin{equation}
h_i \longrightarrow \tilde{h}_i \stackrel{\text{def}}{=} \big(\rho_i,\ \{\phi_k^{(i)}\}_{k=1}^{K_i},\ r_i,\ a_i,\ e_i \big),
\label{eq:case-schema}
\end{equation}
 
\noindent where $\rho_i$ is the reviewer assessment, $\{\phi_k^{(i)}\}_{k=1}^{K_i}$ the chronological timeline, $r_i$ the root cause when established, $a_i$ the documented resolution or mitigation steps, and $e_i$ the troubleshooting-relevant entities. To construct this representation, workers process the ordered artifacts in bounded batches, carrying the evolving case state into each subsequent pass. Each worker receives the next batch alongside the metadata and accumulated state, adding new findings or revising earlier interpretations as evidence develops. Source-query and domain-specific tools provide additional evidence when needed. Once all batches have been processed, a reviewer checks and refines the completed state, consulting source evidence and revision history to resolve omissions or inconsistencies, and produces $\rho_i$. This workflow accommodates histories beyond a single context window while separating incremental extraction from final review. Further details of the extraction workflow are provided in Appendix~\ref{appendix:raft-agent-extraction}.
 
\paragraph{Case assessment and filtering.} The application-specific assessment $\rho_i$ can include case labels, actionability judgments, and supporting reasoning. Filters based on the extracted state (including entities $e_i$), reviewer assessment $\rho_i$, and metadata $m_i$ can be applied during indexing to omit cases from storage or at retrieval time to narrow the search space by error code, category, product version, timestamp, or case outcome. We denote the indices of cases retained in the search index by $\mathcal{I}$.

\paragraph{Timeline entries.} Each timeline entry $\phi^{(i)}_k$ distills a contiguous segment of the case history, comprising consecutive turns $\big(x^{(i)}_j, x^{(i)}_{j+1}, \dots,x^{(i)}_{j+z}\big)$ that together capture a meaningful stage of the investigation. These semantic segments need not coincide with the workflow's processing-batch boundaries. Segmentation follows a single principle: a new
entry begins at each meaningful state transition, where the framing of the problem
or the current understanding is materially updated. These transitions correspond to
the natural phases of an investigation, for example the opening symptom report, a
hypothesis being added, discarded, or confirmed, the root cause being confirmed, and
a resolution being proposed and verified. Acknowledgments and minor updates that add
no new insight are absorbed into the current entry. This keeps the timeline compact ($K_i \ll T_i$) while ensuring every entry carries actionable information. Each entry records the technical state of its segment: the actions taken, the hypotheses under investigation, and the current understanding of the issue.

\paragraph{Case-level graph.} Optionally, we construct an undirected graph $\mathcal{G} = (\mathcal{V}, \mathcal{E})$ over stored cases, with vertex set $\mathcal{V} = \{\tilde{h}_i : i \in \mathcal{I}\}$. The text used to link cases is configurable: root cause, issue summary, or another deployment-specific field can be used alone or in combination. In our experiments, we concatenate root-cause and resolution texts and score every pair of cases using a hybrid of semantic similarity and BM25 lexical similarity~\citep{bm25s}, combined through Reciprocal Rank Fusion (RRF)~\citep{cormack2009reciprocal}. We connect each case to its top-$k$ highest-scoring neighbors, symmetrize the result to obtain $\mathcal{E}$, and assign shared-nearest-neighbor (SNN) weights to the edges. Filters over metadata, entities, and reviewer assessments can further constrain the neighbor set during graph construction. The graph supports principled expansion beyond initial entry-level matches at retrieval time; outside this retrieval path, its communities can also support aggregate analysis of recurring issue families, although we do not evaluate that use here.
 
\subsection{Retrieval}
 
We embed every timeline entry $\phi_k^{(i)}$ across all indexed cases ($i \in \mathcal{I}$). Given a query $q$, we first apply any user-specified case filter, then rank entries from the remaining cases using the same hybrid score (semantic plus lexical via RRF). We greedily promote ranked entries to their parent cases, selecting up to $n$ distinct cases within a predefined context budget. The full procedure is given in Algorithm~\ref{alg:raft-retrieval}.

\begin{algorithm}[t]
\small
\caption{RAFT Entry-Level Retrieval}
\label{alg:raft-retrieval}
\begin{algorithmic}[1]
\REQUIRE Query $q$, indexed timeline entries $\{\phi_k^{(i)}\}$, number of cases $n$, context budget $B$, optional case filter $F$
\ENSURE Retrieval result $R$
\STATE If $F$ is supplied, retain only entries whose parent cases satisfy $F$
\STATE Compute hybrid score $s(q, \phi_k^{(i)})$ for the remaining entries
\STATE Sort entries in descending order of $s$
\STATE Initialize $C \leftarrow \emptyset$, $k^{*} \leftarrow \{\}$, $b \leftarrow 0$
\FOR{each entry $\phi_k^{(i)}$ in sorted order}
    \IF{parent case $i \notin C$}
        \IF{$b + \operatorname{size}(\tilde{h}_i) > B$}
            \STATE \textbf{break}
        \ENDIF
        \STATE $C \leftarrow C \cup \{i\}$; $k^{*}_{i} \leftarrow k$
        \STATE $b \leftarrow b + \operatorname{size}(\tilde{h}_i)$
    \ENDIF
    \IF{$|C| = n$}
        \STATE \textbf{break}
    \ENDIF
\ENDFOR
\STATE \textbf{return} $R = \{(\tilde{h}_c,\ k^{*}_{c}) : c \in C\}$
\end{algorithmic}
\end{algorithm}
 
For each selected case $c \in C$, $k^{*}_{c}$ is the index of the highest-scoring timeline entry, so the agent receives both the full case representation $\tilde{h}_c$ and the anchor entry that triggered the match. This lets the agent see which investigation state was matched, not just which case.

Once the agent has identified seed cases via entry-level retrieval, the case-level graph $\mathcal{G}$ enables expansion to neighbors that are similar under the configured linking view but may not match at the entry level. For example, with the root-cause and resolution view used in our experiments (see Appendix~\ref{appendix:graph}), linked cases may share an underlying cause or remediation strategy despite exhibiting different symptoms or intermediate states. Simply increasing the initial $n$ tends to introduce noise rather than uncover these complementary cases \citep{zhang2025g}. Graph expansion provides a targeted way to retrieve them while keeping the initial retrieval selective.
 
\section{Experiments}
\label{experiments}
 
We evaluate RAFT's retrieval performance against vanilla RAG and GraphRAG baselines on a synthetic development benchmark, and then test whether the results transfer to real cases with human-created labels (Section~\ref{sec:jira}).

\subsection{Synthetic Development Benchmark}
 
As discussed in Section~\ref{sec:related_work}, no existing public dataset meets the requirements for evaluating retrieval in multi-stage troubleshooting. We therefore construct a synthetic corpus grounded in Microsoft Learn Windows Server troubleshooting documentation~\citep{MicrosoftDocs2024SupportArticles}. The synthesis pipeline first organizes source articles into a structured wiki, then generates 2--4 support cases per documented root cause, injecting context from related articles to produce realistic diagnostic ambiguity. For evaluation, we hold out one case per root-cause group as the test query, with the remaining cases forming the indexed corpus. The synthetic data generation pipeline is detailed in Appendix~\ref{appendix:mslearn}; Table~\ref{table:dataset} reports per-category statistics.
 
\begin{table}[t]
  \centering
  \begin{adjustbox}{max width=\columnwidth}
  \begin{tabular}{lrcc}
    \toprule
    \textbf{Category} & \textbf{Cases}
      & \textbf{Messages} & \textbf{Tokens} \\
    \midrule
    Active Directory         & 177 & 11.0\;$\pm$\;1.8 & 3297\;$\pm$\;795 \\
    Windows Security         & 121 & 9.7\;$\pm$\;1.8  & 2556\;$\pm$\;628 \\
    Remote Desktop           & 72  & 10.7\;$\pm$\;1.9 & 2794\;$\pm$\;657 \\
    Group Policy             & 92  & 9.8\;$\pm$\;2.2  & 2948\;$\pm$\;642 \\
    Licensing and Activation & 62  & 11.1\;$\pm$\;1.9 & 2550\;$\pm$\;496 \\
    Networking               & 245 & 10.4\;$\pm$\;1.7 & 2346\;$\pm$\;579 \\
    Backup and Storage       & 57  & 10.4\;$\pm$\;2.2 & 3292\;$\pm$\;897 \\
    \midrule
    Overall                  & 826 & 10.4\;$\pm$\;1.9 & 2767\;$\pm$\;770 \\
    \bottomrule
  \end{tabular}
  \end{adjustbox}
  \caption{Synthetic dataset statistics per category.
    Number of Messages and Tokens report per-case
    mean\;$\pm$\;std.}
  \label{table:dataset}
\end{table}
 
\subsection{Baselines}
 
We compare RAFT against vanilla RAG and two recent GraphRAG methods: HippoRAG2~\citep{gutierrez2025rag} and Fast-GraphRAG~\citep{circlemind2024fastgraphrag}. HippoRAG2 builds an open-relation knowledge graph over the corpus and retrieves via personalized PageRank seeded by query-linked entities; Fast-GraphRAG extracts an entity graph and returns a budgeted mix of entities, relations, and chunks. We run both in their default configurations. To ensure a fair comparison, all methods share the same embedding model (\texttt{text-embedding-3-large}) and the same LLM for indexing (\texttt{gpt-5.2}); evaluation uses \texttt{gpt-5.4}. No metadata filtering is applied, isolating the effect of each method's retrieval mechanism. Moreover, every case in our corpus is actionable, so RAFT's actionability filtering excludes no cases and provides no advantage in this comparison. In production, retrieval tools exposed to an agent typically enforce a per-call token cap on returned context. Real support cases are token-intensive, often running into tens of thousands of tokens, whereas our synthetic cases are considerably shorter, averaging 2767 tokens (Table~\ref{table:dataset}). To reflect this constraint, we cap retrieved context at 6000 tokens for all methods. For methods that return case-level units (RAFT, Vanilla RAG, HippoRAG2), we additionally cap retrieval at 5 distinct cases; Fast-GraphRAG returns entity-linked chunks rather than case-level units, so only the token cap applies. Detailed configurations are provided in Appendix~\ref{appendix:experiments_configs}.
 
\subsection{Evaluation Metrics}
 
A key property of troubleshooting is that useful guidance depends on how far the investigation has progressed. To capture this, we construct queries from prefixes of each test case at three progress points: 0\%, 30\%, and 60\% of turns. The 0\% query contains only the initial symptom report; later cutoffs reveal progressively more diagnostic context. We report three metrics:
 
\begin{itemize}[leftmargin=*, itemsep=3pt]
\item \textbf{Case Hit:} whether at least one retrieved passage belongs to a ground-truth similar case, i.e., one sharing the same root cause and resolution.
\item \textbf{Root Cause Coverage:} the fraction of atomic claims in the gold root-cause explanation that are entailed by the retrieved context, as judged by an LLM.
\item \textbf{Resolution Steps Coverage:} the analogous fraction for the gold remediation procedure.
\end{itemize}
 
Case Hit measures whether retrieval finds a matching case, while coverage measures how much of the gold diagnostic and remediation evidence the retrieved context supports. Full metric definitions appear in Appendix~\ref{appendix:experiments_metrics}.
 
\subsection{Results on the Synthetic Benchmark}
\label{sec:results}

Table~\ref{tab:results_retrieval-performance} reports retrieval performance across all methods and progress levels. RAFT achieves the best scores on every metric, with the largest and most reliable gains on Case Hit. At 0\% progress (initial symptom only), RAFT achieves 84.2\% Case Hit compared to 67.3\% for vanilla RAG and 65.0\% for HippoRAG2. This advantage persists as more context becomes available, with RAFT reaching 88.8\% Case Hit at 60\% progress. To statistically evaluate this claim, we compare RAFT against vanilla RAG, the strongest baseline, using bootstrap resampling clustered by root-cause group; the Case Hit gains are statistically significant at all three progress points, with full confidence intervals in Appendix~\ref{appendix:uncertainty}.
 
Fast-GraphRAG underperforms vanilla RAG at all progress levels. This aligns with recent findings that complex entity extraction and graph-based reasoning offer little benefit, and can even degrade retrieval, when the task does not demand hierarchical knowledge retrieval or deep contextual reasoning across documents~\citep{xiang2025use}. In our setting, surfacing similar cases and supplying actionable insights matter more than abstract relational inference. A similar pattern holds for HippoRAG2, which also fails to surpass vanilla RAG in this setting.
 
To probe \emph{why} entry-level retrieval helps, we examine where within a matched case the hit occurs. For each test case with a correct retrieval, we record the timeline entry that triggered the match and report its absolute index and its depth as a percentile of the number of timeline entries in the extracted case, averaged over hits, in Table~\ref{tab:matched-entry-depth}. The match moves steadily deeper as the query reflects a later stage, from 9.1\% depth at 0\% progress to 54.0\% at 60\%. This is the intended behavior: early queries carry only the symptom and match the opening entries of past cases, whereas later queries align with the corresponding intermediate states rather than re-matching symptoms.

To assess robustness to noisy queries, we perturb test queries with off-topic content, typos, and dropout, and find that RAFT degrades less than vanilla RAG (Appendix~\ref{appendix:robustness}). In Appendix~\ref{appendix:experiments-additional_experiments}, we examine how indexing-model capacity affects RAFT's retrieval performance through an ablation and present an agentic case study in which the agent further improves retrieval quality by composing its own queries and filtering conditions.

\begin{table*}[!ht]
\small
\centering
\begin{adjustbox}{max width=\textwidth}
\begin{tabular}{l|ccc|ccc|ccc}
\toprule
 & \multicolumn{3}{c|}{\textbf{Case Hit}} & \multicolumn{3}{c|}{\textbf{Root Cause Cov.}} & \multicolumn{3}{c}{\textbf{Resolution Steps Cov.}} \\
\cmidrule(lr){2-4}\cmidrule(lr){5-7}\cmidrule(lr){8-10}
\textbf{Method} & \textbf{0\%} & \textbf{30\%} & \textbf{60\%} & \textbf{0\%} & \textbf{30\%} & \textbf{60\%} & \textbf{0\%} & \textbf{30\%} & \textbf{60\%} \\
\midrule
Vanilla RAG          & 0.673 & 0.719 & 0.769 & 0.597 & 0.625 & 0.672 & 0.528 & 0.548 & 0.590 \\
HippoRAG2~\citep{gutierrez2025rag}            & 0.650 & 0.688 & 0.711 & 0.574 & 0.609 & 0.637 & 0.507 & 0.541 & 0.559 \\
Fast-GraphRAG~\citep{circlemind2024fastgraphrag}        & 0.421 & 0.442 & 0.583 & 0.294 & 0.299 & 0.448 & 0.208 & 0.213 & 0.341 \\
\textbf{RAFT} (Ours) & \textbf{0.842} & \textbf{0.871} & \textbf{0.888} & \textbf{0.649} & \textbf{0.675} & \textbf{0.689} & \textbf{0.563} & \textbf{0.587} & \textbf{0.605} \\
\bottomrule
\end{tabular}
\end{adjustbox}
\caption{Retrieval evaluation results averaged over 5 independent runs, reported at three progress points (0\%, 30\%, 60\%). For each run, we randomly shuffle all cases and select one test case per root-cause group; the remaining cases form the indexed corpus. Best result per column in \textbf{bold}.}
\label{tab:results_retrieval-performance}
\end{table*}
 
\begin{table}[t]
\centering
\begin{adjustbox}{max width=\columnwidth}
\begin{tabular}{@{}ccc@{}}
\toprule
Case      & Matched-Entry   & Matched-Entry \\
Progress  & Depth (\%)      & Index         \\
\midrule
\textbf{0\%}   & 9.1        & 0.28 \\
\textbf{30\%}  & 20.0       & 0.59 \\
\textbf{60\%}  & 54.0       & 1.58 \\
\bottomrule
\end{tabular}
\end{adjustbox}
\caption{Position of the retrieval match within a case, averaged over hits.
\emph{Depth} = position as a percentile of the number of timeline entries in the extracted case ($0\%$ = first entry);
\emph{Index} = absolute position. }
\label{tab:matched-entry-depth}
\end{table}

\subsection{Human-Grounded Evaluation on Apache Jira}
\label{sec:jira}

To test whether RAFT transfers to real datasets, we build an evaluation set from public Apache Jira projects, where engineers link duplicate issues in their normal workflow (construction details in Appendix~\ref{appendix:jira}). The final evaluation set contains 30 audited duplicate groups, each evaluated against 570 additional resolved (\emph{Fixed}) issues as distractors, over contributor-written, unredacted issue histories.

\begin{table}[t]
\centering
\begin{adjustbox}{max width=\columnwidth}
\begin{tabular}{l|ccc}
\toprule
 & \multicolumn{3}{c}{\textbf{Case Hit}} \\
\cmidrule(lr){2-4}
\textbf{Method} & \textbf{0\%} & \textbf{30\%} & \textbf{60\%} \\
\midrule
Vanilla RAG          & 0.667 & 0.667 & 0.789 \\
\textbf{RAFT} (Ours) & \textbf{0.833} & \textbf{0.840} & \textbf{0.895} \\
\bottomrule
\end{tabular}
\end{adjustbox}
\caption{Case Hit on the Apache Jira evaluation set (30 human-audited duplicate groups, 570 distractors), averaged over five runs.}
\label{tab:jira}
\end{table}

We apply RAFT as is, with no modifications to the extraction prompt, schema, models, or retrieval procedure from the synthetic experiments, and compare against vanilla RAG, the strongest baseline in Section~\ref{sec:results}. RAFT improves Case Hit by $+16.7$, $+17.3$, and $+10.5$ percentage points at the 0\%, 30\%, and 60\% progress points (Table~\ref{tab:jira}), providing directional evidence that the advantage transfers to real case histories.

\section{Conclusion}
 
We presented RAFT, a stateful retrieval-augmented framework that addresses the limitations of generic RAG and GraphRAG pipelines on enterprise customer support data. RAFT abstracts each closed historical case into a directed chain of timeline entries and optionally connects cases at the graph level through a configurable similarity representation, enabling state-aware retrieval that returns coherent, stage-specific evidence. On a synthetic benchmark built from Microsoft Learn Windows Server documentation, RAFT substantially improves Case Hit over vanilla RAG and recent GraphRAG baselines at every progress level, and an evaluation on real Apache Jira issues provides directional evidence that the advantage transfers to real case histories.
 
\clearpage
 
\section*{Limitations}

While our experiments demonstrate the effectiveness of RAFT for retrieval, three limitations should be noted. First, our main evaluation is conducted on a synthetic dataset of moderate scale, whereas production corpora typically contain far more cases with substantially higher token counts per case. Second, the Apache Jira evaluation comprises 30 audited duplicate groups and carries no confidence intervals, so we treat it as directional transfer evidence rather than a comprehensive real-world evaluation. Third, this work does not evaluate final diagnosis, resolution success, engineer productivity, or other end-to-end troubleshooting outcomes; we scope the work to the retrieval layer for the reasons given in Section~\ref{sec:introduction}.

\bibliography{custom}

\appendix
\newpage

\section{Microsoft Learn Synthetic Dataset}
\label{appendix:mslearn}
 
Microsoft Learn is Microsoft's public documentation site for product guidance, troubleshooting articles, and learning resources. Each troubleshooting article typically documents a specific error or issue, including its symptoms, root cause, and resolution steps. Understanding these articles requires technical knowledge spanning multiple products, concepts, and technologies, which makes them a strong foundation for synthetic support case generation. We construct the dataset in two phases: we first build a \textbf{knowledge base} from the source articles, and then generate synthetic cases grounded in that knowledge base and troubleshooting articles.

\paragraph{Knowledge base construction.} Rather than asking agents to read individual articles in isolation and generate cases from them, we first organize the source material into a wiki-style knowledge base. We focus on Windows Server troubleshooting and select seven categories: Active Directory, Group Policy, Licensing and Activation, Remote Desktop, Windows Security, Backup and Storage, and Networking. We use the Claude Code CLI with \texttt{claude-opus-4-7} to construct this knowledge base, running one session per category. Each session spawns sub-agents that work on individual subcategories; each sub-agent reads all source files within its scope and classifies them as troubleshooting, informational, or general knowledge. The orchestrating agent then consolidates these results and produces a category overview.

\paragraph{Case generation.} From the indexed knowledge base, we generate synthetic cases on a per-article basis, again with one Claude Code session per category. To refine the generation process, we apply a simple self-reflection strategy executed by Claude Code itself. Starting from a high-level generation instruction file in Markdown, the agent produces 15 pilot cases by spawning generation sub-agents, then reviews the resulting cases, identifies their shortcomings, and revises the instruction file accordingly. This loop continues until no meaningful improvement is observed, at which point the full generation session is launched for the category.

\paragraph{Role of the wiki.} The key benefit of generating from the wiki is that agents are not confined to the single article they are working from; they also gain awareness of related articles within the same topic area, including their symptoms, root causes, and points of overlap. This cross-article context is what makes the generated cases diagnostically challenging: agents can introduce realistic red herrings drawn from sibling issues, construct multi-step diagnostic paths in which plausible alternative causes are explored and ruled out, and ensure that no two conversations follow the same troubleshooting sequence. Without the wiki, agents operate in isolation and produce cases that are technically correct but diagnostically straightforward. With the wiki, an initial symptom can plausibly point to several documented root causes, requiring the troubleshooting conversation to disambiguate among them.

\paragraph{Summary.} Using this two-phase approach, first building a structured knowledge base and then generating cases informed by it, we processed over 800 source articles across seven Windows Server categories and produced 826 synthetic support cases. We generate 2--4 cases per root cause so that, during evaluation, each cause group can be split into an indexing set and a test set; all cases within the same group share the same root cause and resolution steps. Below is a short example of a synthetic case.

\begin{tcolorbox}[
    breakable,
    enhanced,
    colback=gray!3,
    colframe=black!70,
    boxrule=0.5pt,
    arc=2pt,
    left=4pt, right=4pt, top=3pt, bottom=3pt,
    fonttitle=\bfseries\small,
    title={Example Case \texttt{TKT-ADAUDIT-05} (Group: SEC --- Permissions, Access Control, and Auditing)}
  ]
  \footnotesize

  \textbf{\textcolor{blue!60!black}{[Customer $\rightarrow$ Support]}}\quad
  \emph{Subject: Auditing logonHours changes for service accounts}\\[2pt]
  We have an OU of service accounts (\texttt{OU=ServiceAccounts,DC=tailspintoys,DC=local}).
  A pen-test finding flagged that someone could remove the \texttt{logonHours}
  restriction on a service account and use it outside maintenance windows. We want
  to log \emph{any} change to \texttt{logonHours} on accounts in that OU.
  \texttt{Directory Service Access} and \texttt{Directory Service Changes} auditing
  are already on. Just need the right SACL to capture \texttt{logonHours} changes
  specifically (not every attribute). Server 2019 DCs.

  \vspace{4pt}\hrule\vspace{4pt}

  \textbf{\textcolor{green!50!black}{[Agent $\rightarrow$ Customer]}}\quad
  \emph{Subject: RE: SACL for logonHours auditing}\\[2pt]
  For narrow attribute-specific auditing you want a SACL targeting just the
  \texttt{logonHours} property write. On \texttt{OU=ServiceAccounts}:
  \begin{itemize}[leftmargin=1.2em,topsep=2pt,itemsep=1pt]
    \item ADUC $>$ View $>$ Advanced Features.
    \item Right-click the OU $>$ Properties $>$ Security $>$ Advanced $>$
          Auditing $>$ Add.
    \item Principal: \texttt{Everyone}; Type: \texttt{All};
          Applies to: \texttt{Descendant User objects}.
    \item Leave \texttt{Write all properties} unchecked; tick only
          \texttt{Write logonHours}.
  \end{itemize}
  You will get a \texttt{5136} event only when \texttt{logonHours} is modified,
  with the old/new byte array---keeping noise down.

  \vspace{4pt}\hrule\vspace{4pt}

  \textbf{\textcolor{blue!60!black}{[Customer $\rightarrow$ Agent]}}\quad
  \emph{Subject: RE: SACL for logonHours auditing}\\[2pt]
  Test passed first try. \texttt{5136} with
  \texttt{AttributeLDAPDisplayName = logonHours} and the new byte value
  (\texttt{OperationType 14674}). Exactly what we wanted. Thanks.

  \vspace{4pt}\hrule\vspace{4pt}

  \textbf{\textcolor{purple!70!black}{Synthetic Annotation}}\\[2pt]
  \textbf{Root cause:} Customer needed narrow attribute-specific auditing of
  \texttt{logonHours} changes under \texttt{OU=ServiceAccounts}; broad SACLs would
  be too noisy, so the SACL had to be scoped to the \texttt{logonHours} property
  only.\\[2pt]
  \textbf{Resolution:} Targeted SACL via ADUC Auditing entry (Principal
  \texttt{Everyone}, Type \texttt{All}, \texttt{Descendant User objects}), ticking
  only \texttt{Write logonHours}. Validated via a \texttt{logonHours} toggle
  producing event \texttt{5136}.

  \end{tcolorbox}

\section{Case Extraction: Output Models and Workflow}
\label{appendix:raft}

\subsection{Output Models}
\label{appendix:raft-schema definitions}

\lstdefinestyle{pydantic}{
    language=Python,
    basicstyle=\footnotesize\ttfamily,
    keywordstyle=\color{blue!70!black}\bfseries,
    commentstyle=\color{gray},
    stringstyle=\color{green!50!black},
    morekeywords={class,bool,str,list,None,Field,BaseModel,Entity,TimelineEntry,min_length,max_length,max_length},
    showstringspaces=false,
    columns=fullflexible,
    breaklines=true,
    frame=single,
    framerule=0.4pt,
    rulecolor=\color{black!60},
    xleftmargin=2pt,
    aboveskip=4pt, belowskip=4pt,
  }

\begin{lstlisting}[style=pydantic,caption={Our output models.},label={lst:interaction-graph}]
from typing import Annotated, Self
from pydantic import BaseModel, ConfigDict, Field, model_validator

class CaseExtraction(BaseModel):

    model_config = ConfigDict(extra="forbid")

    entities: list[Annotated[str, Field(max_length=120)]] = Field(max_length=25)
    timeline: list[Annotated[str, Field(max_length=4800)]]
    root_cause: str | None = Field(max_length=4800)
    resolution_steps: str | None = Field(max_length=4800)


class CaseReview(BaseModel):
    """Final eligibility assessment."""

    model_config = ConfigDict(extra="forbid")

    extractable: bool = Field(strict=True)
    non_extractable_reasoning: str | None

    @model_validator(mode="after")
    def reasoning_matches_assessment(self) -> Self:
        if self.extractable:
            if self.non_extractable_reasoning is not None:
                raise ValueError("An extractable case must have null non_extractable_reasoning")
        elif (
            self.non_extractable_reasoning is None
            or not self.non_extractable_reasoning.strip()
        ):
            raise ValueError("A non-extractable case requires nonblank non_extractable_reasoning")
        return self
  \end{lstlisting}

\subsection{Agent-based Case Extraction}
\label{appendix:raft-agent-extraction}

\begin{figure*}[t]
  \centering
  \includegraphics[width=\textwidth]{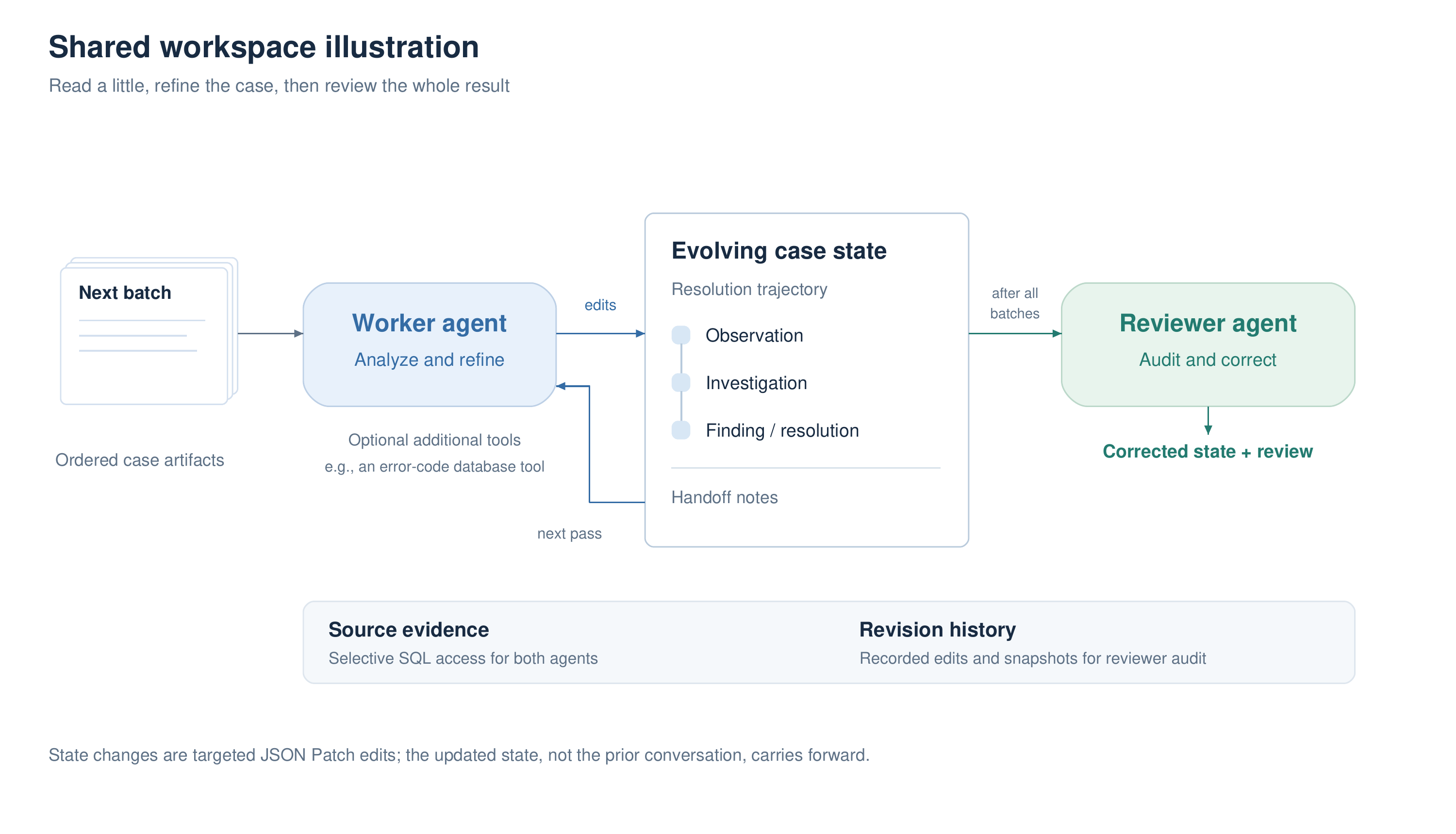}
  \caption{Agent-based case extraction workflow. A worker processes bounded batches and updates an evolving case state, which is carried into subsequent passes. Both agents can selectively query source evidence, and the reviewer can additionally inspect revision history, correct the completed state, and produce a structured assessment. Workers may use additional domain-specific tools. }
  \label{fig:raft-agent-workflow}
\end{figure*}

Our worker--reviewer workflow separates evidence synthesis from final assessment (Figure~\ref{fig:raft-agent-workflow}). Workers accumulate and refine a structured case state across bounded inputs; a reviewer checks the completed result and produces an application-defined assessment. Cases are processed independently. The single-worker configuration used in the experiments is described in Appendix~\ref{appendix:raft-indexing_process}.

\paragraph{Context budget and model choice.} Processing bounded inputs allows case histories to exceed the model's context window without placing the entire history in a single invocation. It is also intended to mitigate the degradation associated with long inputs~\citep{hong2025context}. The batch size can be adjusted to the model's context capacity and ability to reliably synthesize complex evidence, leaving room for instructions, metadata, the accumulated state, and tool interactions. 

\paragraph{Bounded, incremental processing.} The runner partitions ordered artifacts using a configurable character/token threshold on the source content supplied to each worker. Each invocation receives the next batch, case metadata, and the current state in a fresh conversational context. These processing batches are distinct from semantic timeline segments: a pass may produce multiple entries or revise earlier entries, and a timeline segment can span batch boundaries.

\paragraph{An editable state with selective evidence access.} Workers update the shared case state through JSON Patch, which expresses targeted additions, replacements, and removals. This permits later evidence to correct earlier interpretations without requiring the entire output to be regenerated. The state also includes handoff notes for unresolved questions or context that subsequent passes should revisit. All source artifacts remain accessible through a read-only SQLite query tool, allowing agents to inspect relevant records or selected fields on demand. Workers may additionally use user-provided tools; for example, an error-code database can provide domain knowledge when normalizing the entities in $e_i$.

\paragraph{Final review and traceability.} Successful state commits record snapshots and edits in a revision history. After all batches have been processed, the reviewer receives the completed state and case metadata. It can selectively query source evidence and revision history to investigate omissions or conflicting interpretations, correct the state using the same editing mechanism, and return a separate structured assessment $\rho_i$.

\paragraph{Assessment and downstream policy.} The assessment schema is application-defined. It can contain case-type or outcome labels, such as \emph{mitigated}, \emph{request for information (RFI)}, or \emph{resolved}, alongside actionability judgments and supporting reasoning. Labels characterize a case rather than automatically excluding it. The workflow returns the complete state and assessment; user-defined filters can then determine which cases to store or which stored cases to include in retrieval ranking. The actionability flag in Figure~\ref{fig:raft-overview} illustrates one simple use of this more general assessment.

\subsection{Indexing Process}
\label{appendix:raft-indexing_process}

Cases in our experimental datasets average only a few thousand tokens, so each case is processed by a single worker invocation.

\section{Experiments}
 
\subsection{Metrics}
\label{appendix:experiments_metrics}
 
We evaluate retrieval quality with three complementary metrics. Let
$\mathcal{D}$ denote the evaluation set. Each example
$i \in \mathcal{D}$ consists of a query $q_i$, a ground-truth
\emph{shared id} $s_i^\star$ identifying the underlying
issue, gold root-cause text $r_i^\star$, gold resolution-steps
text $a_i^\star$, the retrieved context string $C_i$, and
the set of retrieved case ids $\mathcal{R}_i$. We write
$\mathrm{sid}(c)$ for the shared id of a retrieved case $c$.

\paragraph{Case Hit.}
Measures whether retrieval surfaces \emph{any} case
belonging to the same underlying issue as the query. 
\begin{equation}
\mathrm{CH}_i = \mathds{1} \!\left[\, s_i^\star \in
\{\mathrm{sid}(c) : c \in \mathcal{R}_i\} \,\right],
\end{equation}
\begin{equation}
\mathrm{CH} = \frac{1}{|\mathcal{D}|}
\sum_{i \in \mathcal{D}} \mathrm{CH}_i.
\end{equation}

\paragraph{Root Cause Coverage.}
Quantifies how much of the gold root-cause explanation is
actually \emph{supported} by the retrieved context. An LLM
judge (i)~atomizes $r_i^\star$ into a set of claims
$\mathcal{K}(r_i^\star) = \{k_1, \dots, k_{n_i}\}$, and
(ii)~verdicts each claim $k_j$ against $C_i$ as
$v_{ij} \in \{0,1\}$, where $v_{ij}=1$ iff $k_j$ is entailed
by $C_i$:
\begin{equation}
\mathrm{RCC}_i = \frac{1}{n_i}\sum_{j=1}^{n_i} v_{ij},
\end{equation}
\begin{equation}
\mathrm{RCC} = \frac{1}{|\mathcal{D}|}
\sum_{i \in \mathcal{D}} \mathrm{RCC}_i.
\end{equation}
This rewards retrieving evidence that actually diagnoses the
issue, rather than merely co-occurring with the correct case.

\paragraph{Resolution Steps Coverage.}
Analogously measures how much of the gold remediation
procedure is supported by the retrieved context. The gold
steps $a_i^\star$ are atomized into
$\mathcal{K}(a_i^\star) = \{k_1, \dots, k_{m_i}\}$ and each
claim is verdicted against $C_i$ to obtain
$u_{ij} \in \{0,1\}$:
\begin{equation}
\mathrm{RSC}_i = \frac{1}{m_i}\sum_{j=1}^{m_i} u_{ij},
\end{equation}
\begin{equation}
\mathrm{RSC} = \frac{1}{|\mathcal{D}|}
\sum_{i \in \mathcal{D}} \mathrm{RSC}_i.
\end{equation}
This rewards retrieving the evidence needed to
\emph{act on} the issue.

\subsection{Baseline Configurations}
\label{appendix:experiments_configs}

\begin{itemize}
\item \textbf{Vanilla RAG}: We use a chunk size of 256 tokens with a 64-token overlap for indexing. At retrieval time, we return the unique cases associated with the top-ranked chunks, in rank order.
\item \textbf{HippoRAG2}: We use the default settings and operate directly on each case. Their implementation applies no chunking by default; this suffices because all cases fit within the token limit of our chosen embedding model, and preliminary experiments with a chunked variant yielded worse performance for this baseline.
\item \textbf{Fast-GraphRAG}: We adopt the default chunking configuration from the implementation. For retrieval, we preserve the implementation's default budget ratios and apply them to our 6{,}000-token budget, allocating 1{,}500 tokens to entities, 1{,}125 tokens to relations, and 3{,}375 tokens to chunks.
\end{itemize}

\subsection{Uncertainty Analysis}
\label{appendix:uncertainty}

For the paired uncertainty analysis, we rerun RAFT and vanilla RAG on identical held-out groups, progress points, and context budgets across the five splits, producing a paired result for every query. We then bootstrap the underlying issue-group clusters across all splits, rather than treating only five split-level averages as the inferential sample, and report percentile 95\% confidence intervals for the paired differences in Table~\ref{tab:uncertainty}. All three Case Hit gains are statistically supported. Every unconditional coverage point estimate also favors RAFT; the early-stage coverage gains and the 30\% root-cause gain are statistically supported, the 30\% resolution gain is small and near the interval boundary, and we make no late-stage coverage claim.

\begin{table}[!ht]
\centering
\begin{adjustbox}{max width=\columnwidth}
\begin{tabular}{llr}
\toprule
\textbf{Metric} & \textbf{Progress} & \textbf{RAFT $-$ Vanilla (pp) [95\% CI]} \\
\midrule
Case Hit   & 0\%  & $+16.79$ [$+13.91$, $+19.78$] \\
Case Hit   & 30\% & $+14.79$ [$+12.02$, $+17.73$] \\
Case Hit   & 60\% & $+12.19$ [$+9.75$, $+14.68$] \\
\midrule
Root cause & 0\%  & $+5.69$ [$+3.35$, $+8.07$] \\
Root cause & 30\% & $+4.15$ [$+1.88$, $+6.56$] \\
Root cause & 60\% & $+1.88$ [$-0.07$, $+3.89$] \\
\midrule
Resolution & 0\%  & $+2.98$ [$+1.03$, $+4.99$] \\
Resolution & 30\% & $+2.01$ [$+0.09$, $+3.97$] \\
Resolution & 60\% & $+0.57$ [$-1.18$, $+2.32$] \\
\bottomrule
\end{tabular}
\end{adjustbox}
\caption{Paired differences between RAFT and vanilla RAG in percentage points (pp), with issue-group-clustered bootstrap 95\% CIs.}
\label{tab:uncertainty}
\end{table}

\subsection{Query Robustness}
\label{appendix:robustness}

We run a paired controlled-noise study at 30\% and 60\% progress across the same five held-out splits. For each clean query, both RAFT and vanilla RAG receive the identical deterministic perturbation: (i) adjacent-character swaps in 2\% of words of length at least five, (ii) removal of 40\% of observed updates while preserving the initial report, or (iii) insertion of one turn from an unrelated issue category. We then measure each method's Case Hit change relative to its result on the corresponding clean query; Table~\ref{tab:robustness} reports the changes in percentage points, with positive paired differences favoring RAFT.

\begin{table}[!ht]
\centering
\begin{adjustbox}{max width=\columnwidth}
\begin{tabular}{lrrrr}
\toprule
\textbf{Perturbation} & \textbf{Progress} & \textbf{RAFT (pp)} & \textbf{Vanilla (pp)} & \textbf{Paired diff.\ (pp) [95\% CI]} \\
\midrule
Light typos        & 30\% & $-0.11$  & $-0.22$  & $+0.11$ [$-0.44$, $+0.66$] \\
Light typos        & 60\% & $0.00$   & $+0.33$  & $-0.33$ [$-1.05$, $+0.39$] \\
40\% update dropout & 30\% & $-0.39$  & $-1.05$  & $+0.66$ [$-0.89$, $+2.22$] \\
40\% update dropout & 60\% & $-0.83$  & $-1.27$  & $+0.44$ [$-1.39$, $+2.33$] \\
One unrelated turn & 30\% & $-11.14$ & $-25.93$ & $+14.79$ [$+11.52$, $+18.06$] \\
One unrelated turn & 60\% & $-9.36$  & $-49.31$ & $+39.94$ [$+35.73$, $+44.10$] \\
\bottomrule
\end{tabular}
\end{adjustbox}
\caption{Case Hit change under paired, deterministic query perturbations, in percentage points (pp) relative to the corresponding clean query. Positive paired differences favor RAFT.}
\label{tab:robustness}
\end{table}

The changes under typos and missing updates are small for both methods, and the between-method differences are not statistically resolved. The unrelated turn is substantially harder and degrades vanilla RAG much more than RAFT. These are controlled query-robustness and within-benchmark results, not evidence of noisy-corpus robustness, production-scale behavior, or cross-domain generalization.

\subsection{Graph Contribution and Sensitivity}
\label{appendix:graph}

In graph construction, $k$ is a pre-symmetrization sparsity cap: we rank cross-case candidates using the hybrid semantic/lexical score, retain the top $k$, remove links below a fixed 0.6 embedding-similarity threshold, and then symmetrize the retained links, so a case can acquire more than $k$ final neighbors through incoming links.

To isolate the graph's marginal contribution, we compare pure entry-level retrieval with a budget-matched graph condition that reserves one of the five case slots for one-hop expansion from the directly retrieved seeds. \emph{Full-sibling recovery} asks, for issue groups with two correct sibling cases available in the index, whether both are retrieved. Table~\ref{tab:graph} reports the sparsest tested setting, $k=3$.

\begin{table}[!ht]
\centering
\begin{adjustbox}{max width=\columnwidth}
\begin{tabular}{rrrr}
\toprule
\textbf{Progress} & \textbf{Direct recovery} & \textbf{With expansion} & \textbf{Change (pp)} \\
\midrule
0\%  & 73.71\% & 75.65\% & $+1.94$ \\
30\% & 79.84\% & 82.74\% & $+2.90$ \\
60\% & 88.71\% & 88.55\% & $-0.16$ \\
\bottomrule
\end{tabular}
\end{adjustbox}
\caption{Full-sibling recovery without and with budget-matched graph expansion ($k=3$; four direct anchors plus one graph-expanded slot).}
\label{tab:graph}
\end{table}

At $k=3$, graph expansion modestly improves sibling diversity at early and intermediate progress; at 60\%, direct retrieval is already high and the result is effectively unchanged. Overall Case Hit changes by only $+0.22$, $-0.28$, and $-0.06$ pp at 0\%, 30\%, and 60\%, respectively. As a sensitivity check, $k=5$ and $k=10$ produce nearly identical results: across the three graph settings, Case Hit varies by at most 0.17 pp and full-sibling recovery by at most 0.81 pp. The conclusion is therefore not sensitive to $k$ in this benchmark, and we use $k=3$ throughout, presenting graph expansion as an optional evidence-diversification mechanism rather than the source of RAFT's main retrieval gain.

\subsection{Additional Experiments}
\label{appendix:experiments-additional_experiments}
 
We conduct two additional experiments. First, we examine how indexing-model capacity affects RAFT's retrieval quality. Second, we evaluate an agentic setting in which the agent composes its own search queries and filtering conditions given the same 0\%, 30\%, and 60\% case context used in the main experiments.
 
Note that in this paper we extract general-purpose entities. For domain-specific deployments, we recommend replacing these with application-specific entities (e.g., error codes, Windows version) that uniquely characterize a case; such entities enable agents to write more targeted queries and further narrow the search space.
 
Table~\ref{tab:ablation} reports an ablation on indexing-model capacity, in which we replace the default indexing model \texttt{gpt-5.2} with the smaller \texttt{gpt-5.4-mini} and \texttt{gpt-5.4-nano}, and separately reduce the reasoning effort from \texttt{medium} to \texttt{low}. Both \texttt{gpt-5.4-mini} and the low-reasoning setting yield only modest drops, whereas \texttt{gpt-5.4-nano} degrades more noticeably, as expected for a substantially less capable model. Extraction quality therefore matters, but smaller models remain viable under cost constraints, offering a practical trade-off for large-scale deployments.
 
\begin{table}[!ht]
\centering
\begin{adjustbox}{max width=\columnwidth}
\begin{tabular}{l|c|c|c}
\toprule
\textbf{Method} & \textbf{Case Hit} & \textbf{Root Cause Coverage} & \textbf{Resolution Steps Coverage} \\
\midrule
\multicolumn{4}{c}{\textbf{0\% Case Progress}} \\
\midrule
RAFT (low reasoning)                  & 0.827 & 0.649 & 0.573 \\
RAFT (\texttt{gpt-5.4-mini})          & 0.823 & 0.603 & 0.536 \\
RAFT (\texttt{gpt-5.4-nano})          & 0.772 & 0.590 & 0.492 \\
\midrule
\multicolumn{4}{c}{\textbf{30\% Case Progress}} \\
\midrule
RAFT (low reasoning)                   & 0.868 & 0.686 & 0.597 \\
RAFT (\texttt{gpt-5.4-mini})           & 0.855 & 0.627 & 0.555 \\
RAFT (\texttt{gpt-5.4-nano})           & 0.800 & 0.611 & 0.545 \\
\midrule
\multicolumn{4}{c}{\textbf{60\% Case Progress}} \\
\midrule
RAFT (low reasoning)                   & 0.886 & 0.705 & 0.629 \\
RAFT (\texttt{gpt-5.4-mini})           & 0.871 & 0.643 & 0.575 \\
RAFT (\texttt{gpt-5.4-nano})           & 0.855 & 0.631 & 0.554 \\
\bottomrule
\end{tabular}
\end{adjustbox}
\caption{Ablation on indexing-model capacity. We replace the default \texttt{gpt-5.2} with smaller models (\texttt{gpt-5.4-mini}, \texttt{gpt-5.4-nano}) to assess sensitivity to extraction quality.}
\label{tab:ablation}
\end{table}
 
In the agentic setting, we expose RAFT to the agent as a tool that filters cases by the category of the active case and retrieves similar cases from agent-supplied search queries. The agent is instructed to compose queries that reflect its current understanding of the issue, using the same case context at the 0\%, 30\%, and 60\% progress points as in the main experiments. As shown in Table~\ref{tab:raft-agent}, this improves Case Hit across all progress points.

\begin{table}[!ht]
\centering
\begin{adjustbox}{max width=\columnwidth}
\begin{tabular}{l|ccc}
\toprule
\textbf{Case Hit} & \multicolumn{3}{c}{\textbf{Case Progress}} \\
\cmidrule(lr){2-4}
\textbf{Method} & \textbf{0\%} & \textbf{30\%} & \textbf{60\%} \\
\midrule
RAFT-Agent & 0.910 & 0.910 & 0.919 \\
\bottomrule
\end{tabular}
\end{adjustbox}
\caption{Agentic retrieval with RAFT. The agent composes its own queries and filtering conditions; Case Hit improves over the non-agentic baseline (Table~\ref{tab:results_retrieval-performance}).}
\label{tab:raft-agent}
\end{table}

\subsection{Deployment Considerations}
\label{appendix:deployment}

Compared with vanilla RAG, RAFT incurs additional LLM cost to distill historical cases during indexing. This reflects a trade-off between offline preparation and inference-time context consumption. Vanilla RAG avoids LLM-based extraction at indexing time, but returning raw cases requires the troubleshooting agent to read long case histories to reconstruct the relevant investigation and resolution, potentially repeating this work across queries. RAFT performs this distillation offline and reuses compact case representations, reducing the tokens needed to convey each case's trajectory. Our retrieval results also show that these representations identify similar cases more accurately under the same context budget (Table~\ref{tab:results_retrieval-performance}). The resulting end-to-end cost trade-off depends on query volume and how much retrieved context the agent consumes.

RAFT also supports incremental updates through independent case processing. Adding or revising a case requires extracting that case and updating the retrieval index and case-level graph, without re-extracting other cases. This independence simplifies maintenance relative to GraphRAG pipelines whose shared entity graphs and community summaries introduce dependencies across documents.

Finally, the extraction workflow summarizes a documented investigation rather than solving the case anew, allowing smaller, cheaper models to perform this offline step. Our indexing-model ablation supports this option: \texttt{gpt-5.4-mini} retains strong retrieval performance, while the larger degradation with \texttt{gpt-5.4-nano} illustrates the trade-off between extraction-model capacity and retrieval quality (Table~\ref{tab:ablation}).

\section{Apache Jira Evaluation Set}
\label{appendix:jira}

We freeze public issue histories from Apache Cassandra, Hadoop, HBase, and Spark. Jira contributors mark a later report as a \emph{Duplicate} of an older issue in their normal workflow; we use the later report as the held-out query and the older issue as the exact target only if it was already resolved \emph{Fixed} before the query opened. Other cases serve as distractors, not certified semantic negatives, because Jira links may be incomplete.

The construction pipeline proceeds as follows. Starting from 2{,}082 raw duplicate pairs, metadata filters retain 85 pairs whose older target was already \emph{Fixed} and whose histories were sufficiently substantive. Pre-disclosure, leakage, length, near-copy, and one-pair-per-component filters reduce these to 41 candidate groups. Manual audit then removes 8 semantically mismatched or ambiguous links and 3 leakage-affected groups, leaving 30 audited groups. The final corpus contains 600 cases: the 30 gold targets and 570 \emph{Fixed} distractors. By project, the 30 groups comprise 12 from Cassandra, 5 from Hadoop, 2 from HBase, and 11 from Spark; the distractors comprise 143 from Cassandra, 143 from Hadoop, 142 from HBase, and 142 from Spark. All 30 groups are evaluable at the 0\% and 30\% progress points; at 60\%, the 19 groups with sufficient pre-disclosure history (6 Cassandra, 4 Hadoop, 2 HBase, 7 Spark) are evaluated.

We apply the same extraction prompt, schema, and retrieval procedure as the synthetic experiments; the one configuration difference is a 5{,}000-token context cap instead of 6{,}000. Each query retrieves five cases, and results are averaged over five indexing seeds. We report Case Hit only, because the dataset provides no gold root-cause or resolution-step annotations from which to compute coverage metrics. Given the 30-group scale, we report no confidence intervals and treat the result as directional transfer evidence.

\end{document}